\documentclass[10pt,twocolumn,letterpaper]{article}

\usepackage{times}
\usepackage{epsfig}
\usepackage{graphicx}
\usepackage{amsmath}
\usepackage{amssymb}
\usepackage{csquotes}
\usepackage{float}
\usepackage{tikz}
\usepackage{pgfplots}
\usepackage{multirow}
\usepackage{rotating}
\usepackage{array}
\usepackage[final]{pdfpages}
\usepackage[noend]{algpseudocode}
\usepackage{algorithm}
\usepackage{caption}
\usepackage{subcaption}

\usepackage{import}
\usetikzlibrary{quotes,arrows.meta}
\usetikzlibrary{positioning}

\def\edgecolor{rgb:blue,4;red,1;green,4;black,3}
\newcommand{\midarrow}{\tikz \draw[-Stealth,line width =0.8mm,draw=\edgecolor] (-0.3,0) -- ++(0.3,0);}

\usepackage{Ball}
\usepackage{Box}
\usepackage{RightBandedBox}

\usetikzlibrary{positioning}
\usetikzlibrary{3d} 

\def\ConvColor{rgb:yellow,5;red,2.5;white,5}

\def\PoolColor{rgb:red,1;black,0.3}

\def\SoftmaxColor{rgb:magenta,5;black,7}

\usepackage[pagebackref=true,breaklinks=true,letterpaper=true,colorlinks,bookmarks=false]{hyperref}

\begin{document}

\title{Genetic Deep Learning for Lung Cancer Screening}

\author{Hunter Park \\
Innovation Dx Inc.\\
{\tt\small hpark@innovationdx.com}
\and
Connor Monahan \\
Innovation Dx Inc. \\
{\tt\small cmonahan@innovationdx.com}
}

\maketitle

\begin{abstract}
Convolutional neural networks (CNNs) have shown great promise in improving computer aided detection (CADe). From classifying tumors found via mammography as benign or malignant to automated detection of colorectal polyps in CT colonography, these advances have helped reduce the need for further evaluation with invasive testing and prevent errors from missed diagnoses by acting as a second observer in today's fast paced and high volume clinical environment. CADe methods have become faster and more precise thanks to innovations in deep learning over the past several years. With advancements such as the inception module and utilization of residual connections, the approach to designing CNN architectures has become an art. It is customary to use proven models and fine tune them for particular tasks given a dataset, often requiring tedious work. We investigated using a genetic algorithm (GA) to conduct a neural architectural search (NAS) to generate a novel CNN architecture to find early stage lung cancer in chest x-rays (CXR). Using a dataset of over twelve thousand biopsy proven cases of lung cancer, the trained classification model achieved an accuracy of 97.15\% with a PPV of 99.88\% and a NPV of 94.81\%, beating models such as Inception-V3 and ResNet-152 while simultaneously reducing the number of parameters a factor of 4 and 14, respectively. 
\end{abstract}

\section{Introduction}

\begin{figure}[!b]
    \centering
    \begin{subfigure}[b]{0.2\textwidth}
        \centering
        \includegraphics[width=\textwidth]{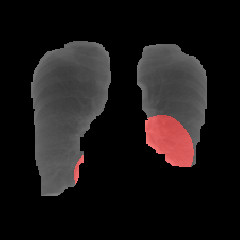}
    \end{subfigure}
    \begin{subfigure}[b]{0.2\textwidth}
        \centering
        \includegraphics[width=\textwidth]{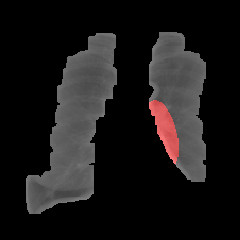}
    \end{subfigure}
    \hfill
    \begin{subfigure}[b]{0.2\textwidth}
        \centering
        \includegraphics[width=\textwidth]{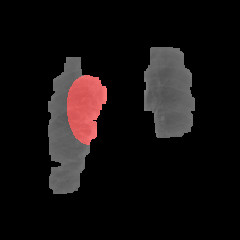}
    \end{subfigure}
    \begin{subfigure}[b]{0.2\textwidth}
        \centering
        \includegraphics[width=\textwidth]{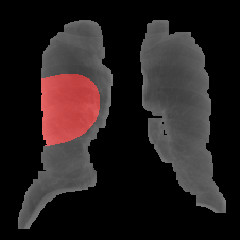}
    \end{subfigure}
    \caption{Sample of output images by the DeepNEAT-Dx}
    \label{fig:sample}
\end{figure}
The rate of lung cancers among the population has dramatically increased during the 20\textsuperscript{th} century, posing a large threat to the public especially those with higher risks to the disease, such as former or current smokers and those with exposure to radiation or chemicals in the workplace. It is estimated that over 1,465,000 people die every year from cancers, 18.2\% of which are a variant of lung cancer \cite{lungcancer}. It is also estimated that there will be 228,150 new diagnoses of lung cancer in the United States in 2019. While the mortality rate has seen steady improvement the past few years, the low survival rate for lung cancer in 2019 of approximately 19\% is attributed to its frequent late detection \cite{lungcancersurvey}. Despite the possibility of detecting lung cancer from annual chest x-rays, such detection in early stages is improbable as the 5-10mm sized lesions are easily overlooked \cite{overlooked}.

Computer aided detection has become a major research topic in medical imaging \cite{cad-survey}. CADe systems provide a ``second opinion" for radiologists to assist them in making a final decision, noting some abnormality that may have been overlooked or mistaken. These systems have become a key component in the detection of breast cancer at a number of screening sites and hospitals \cite{cadbreastcancer}, and many PACS environments are starting to have various CADe programs built into their platforms which reduces the fiscal and technical barriers to use. They are often built on a wide variety of computer vision tools. These systems are able to process many types of medical images including X-rays, MRI, and CT scans, and are intended to support the professional in their practice.

In the paper, we outline the utilization of genetic algorithms to find optimal deep neural network architectures which produces an architecture that beats state of the art models when applied to early detection of lung cancer in CXRs while simultaneously reducing the number of parameters by several orders of magnitude. We then employ a visualization technique to extract a heat map of the activations for the generated model, allowing a radiologist to understand which regions of the CXR lead to an image's classification.

\section{Background}

Over the past several years, image classification and object detection have seen tremendous improvements in accuracy as well as speed thanks to continued research into deep learning and convolutional neural networks (CNNs) \cite{deepcovnets, efficientnet}. The success and improvement found from deep learning is not from improved hardware and more encompassing datasets, but from innovations into model structure. From convolutions into fully connected layers \cite{yann1998}, to the addition of dropout layers \cite{dropout2012}, to optimization \cite{adam, learning2learn}, the approach to deep learning is constantly changing and improving.

\subsection{The Use of CNNs in Medical Applications}

Using image analysis on medical images is not new to doctors or computer scientists. However, many applications of computer vision for medical applications struggle due to medical data being noisy, inexact, sparse, or requiring large amounts of storage.

In a project recently performed at the University of Bern, a group of researchers trained a CNN to classify lung diseases based on lung slices from computed tomography images with an accuracy of 85.5\% on its data set \cite{cnnlungcancer}. They were able to successfully classify lung CT image patches between 6 different lung diseases. Originally, their dataset consisted of 2,032 different diseases. Extremely small training classes creates a major issue in applying deep learning to medical tasks, where rare diseases occur in less than one in ten thousand cases.
To work around this issue, they used a dynamic tree taxonomy to balance their classes and generated classes based on the number of examples rather than final diagnoses. This resulted in only 757 classes instead of 2,032.

In another project performed at the Federal University of Parana, a CNN was employed for classifying images of cell slides of breast cancer patients \cite{breastcancercnns}. Using the BreaKHis dataset \cite{breakhis}, they trained the AlexNet model \cite{alexnet} to classify microscopic biopsy images of benign and malignant breast tumors. Each slide of breast tissue contains 4 images, each with different levels of magnification. To handle the high resolution nature of their dataset, they invoked a variety of computer vision techniques. The first was the use of sliding windows with 50\% overlap and the second was random crops of the raw image with no overlap. Interestingly, the technique was not superior in the task of classification for every level of magnification of the biopsy. 

In \cite{skincancer}, they achieved human specialist performance in several classification tasks dealing with various skin diseases. Their model was capable of identifying benign or malignant skin diseases and distinguishing high level taxonomic classes to identify lesions with similar treatment paths. In addition to this, they also achieved human level capability in specifically discriminating between treatment paths for lesions: biopsy/treat or not. For their training, they used transfer learning from models trained on the ImageNet dataset, despite the image space of ImageNet and their dataset being extremely different. This use of transfer learning may prove to be very effective for other medical tasks.

\subsection{Genetic Algorithms}
Genetic algorithms (GA) were first introduced by John Holland in the early 1970s. Such algorithms are search heuristics that try to mimic the process of natural selection for the purpose of finding possible solutions to optimization problems constructed as search problems. GAs have two main components, the first being the ability to represent every solution in the solution space. In our case, that requires the ability to form every possible CNN architecture, which is achieved through our genetic encoding scheme that is detailed in section 3. The second aspect of GAs is the ability to evaluate the fitness of solutions. In our application this is achieved by first training on the train set and then testing the trained solution on our development set, which then becomes the solution's fitness.

GAs begin with an initial population of genes and a population of genomes that encode a solution to a problem. Once every genome in the population has been evaluated on how well they solve the problem, they are selected based on their fitness to create a new population (generation) of individuals (solutions). Just as in nature, the more fit an individual, the more likely it is it have offspring. As generations are evaluated, the population's fitness \textit{should} trend upwards as each individual gets better at solving the problem, with the hope that at least one individual will be close to optimal parameters to the optimization problem they were designed to solve. 

Evolutionary and genetic algorithms have experienced a resurgence in recent years due to their ability to search irregular and poorly characterized spaces effectively \cite{NAS}, their ability to optimize without expensive gradient computations \cite{GA-RL}, and their ability to be massively distributed \cite{parallelGA}.

\subsection{Neural Architecture Search}
The success of CNNs in perceptual tasks comes from the automation of feature extraction, with features extractors being learned in an end-to-end fashion as opposed to being manually designed. Convolutional neural networks date back to 1989 with LeNet-5 \cite{yann1998}, which was designed to be able to handle variance among identically labeled data by using \enquote{local receptive fields, shared weights, and spatial sub-sampling}. In 2012, AlexNet \cite{alexnet} outlined how CNNs were used to win the ImageNet Large-Scale Visual Recognition Challenge (ILSVRC) in 2012. After AlexNet, the designs trends of CNNs in literature was making them linear and deep. There have been two major breakthroughs in CNN architecture design since AlexNet, which we discuss. 

The first innovation was the inception module \cite{googlenet2015}. This module was designed with two ideas in mind: the approximation of a sparse structure with spatially repeated dense components and the need to keep the computational complexity in check. Unsurprisingly, the inception module does both of these exceptionally well: Inception-V1 used twelve times less parameters while still outperforming AlexNet. Inception architectures take advantage of the fact that convolutional filters of different sizes may extract features from separate clusters of information, and when one repeats them spatially they approximate an optimal sparse structure via dense components. The stacking of different sized convolutional filters concurrently allows for a significant increase in the number of units without an uncontrolled blow-up in computational complexity. This use of dimension reduction allows for shielding the large number of input filters of the last stage to the next layer. Inception architectures follows the principle that visual information should be processed at various scales and then aggregated so that the next stage can extract features from different scales simultaneously. The end result of the inception module is that they are up to three times faster than similarly performing networks with non-inception architecture.

The other major advancement in architecture design came from ILSVRC 2015's winner, ResNet-152 \cite{microsoft2016}. They asked themselves \textit{\enquote{Is learning better networks as easy as stacking more layers?}} They discovered that when deeper networks begin converging, the network's accuracy becomes saturated, and then degrades rapidly. They wanted to ensure that a deeper model should produce no higher training error than its shallower counterpart, so to ensure this, they added identity mappings because solvers often struggle approximating identity mappings by multiple nonlinear layers. Residual learning ensures that solvers can easily approximate identity mappings by simply driving the weights of the multiple nonlinear layers toward zero to approach identity mappings.

Even with these breakthroughs in architecture designs, it is still time consuming to hand design models. Historically, architecture search routines focused on fine-tuning a particular aspect of an architecture \cite{rnnevolve}, however there has been a growing interest in automating the entire architecture design process \cite{prog-nas, eff-nas, diff-nas}. When comparing NAS methods, it is important to not only look at how well of an architecture it produces in terms of accuracy, but also other metrics such as how efficient the search is, the number of parameters of the final architecture, and how scalable the search can be implemented. 

There are two approaches to NAS: heuristic searches \cite{prog-nas, NEAT}, and reinforcement learning (RL) \cite{rl-nas1, rl-nas2}. While both produce similarly impressive models on challenging benchmark datasets, heuristic searches are vastly superior in search efficiency. Due to this, we constructed our optimization problem as a genetic algorithm. 

\begin{algorithm}[!b]
\caption{NEAT Algorithm}
\label{alg:neat}
\begin{algorithmic}

\Procedure{Mutate}{Genome}
\State Initialize Mutation-Rates
\For{each mutation \textbf{in} Mutation-Rates}
    \If{$\epsilon \leq$ mutation.rate}
        \State Apply mutation to Genome
    \EndIf
\EndFor
\State \Return genome
\EndProcedure

\Procedure{New-Generation}{P}
\State \Call{Cull-Species}{P}
\State \Call{Remove-Stale-Species}{P}
\State \Call{Remove-Weak-Species}{P}
\State $P^{'} \gets \varnothing$
\While{\Call{SizeOf}{$P^{'}$} $<$ \Call{SizeOf}{$P$}}
\State $G \gets$ \Call{Random}{$P$.Genomes}
\State $G \gets$ \Call{Mutate}{G}
\State $P^{'} = P^{'} + G$
\EndWhile
\State \Return $P^{'}$
\EndProcedure

\Procedure{NEAT}{Mutation-Rates, Target}
\State Initial-Architecture $\gets$ a minimum NN architecture
\State Best-Genome $\gets \varnothing$
\State $P \gets N$ instances of Initial-Architecture
\While{$\max(P.\mathrm{fitness}) < $ target}
    \State $P \gets$ \Call{New-Generation}{$P$, Mutation-Rates}
    \State Evaluate Population $P$
    \If{$\max(P.\mathrm{fitness}) >$ Best-Genome.fitness}
        \State Best-Genome $\gets P$.Best-Genome
    \EndIf
\EndWhile
\State \Return Best-Genome
\EndProcedure
\end{algorithmic}
\end{algorithm}

\section{Genetic Deep Learning}
There have been many proposed methods detailed how to implement Topology and Weight Evolving Artificial Neural Networks (TWEANNs). \cite{rnnevolve, nngenalg, dualrep}. However, in deep learning there are often times millions of parameters that require optimization, and doing so through any chance based search algorithm is simply impractical. For this paper, we modified the NEAT algorithm as detailed in Algorithm \ref{alg:neat} to evolve the architecture of CNNs, dubbed ``DeepNEAT", which we call DeepNEAT-Dx in alignment to the foundation paper that proposed using NEAT to create CNN architectures \cite{deepneat}. More specifically, we injected convolution and pooling layers with pseudorandom hyperparameters into a minimal convolutional architecture and then optimized the weights through backpropagation on the training set. The fitness of a model is the final accuracy on the development set after 5 epochs of training.

A very critical topic when designing a genetic algorithm is what encoding scheme one should use. This has been an ongoing topic of discussion as encoding schemes directly contribute to the success of genetic algorithms \cite{encoding2}. The requirements for our problem were as follows: the encoding must able to encode directed acyclic graphs (DAGs) of variable size, be able to easily create new, coherent, genomes from the genes of two parent genomes, and lastly the encoding scheme must inherently allow for efficient search for optimal network architectures. With these requirements, there are only a few encoding schemes that are able to meet all these requirements. Following \cite{NEAT}, we used a direct graph encoding scheme, and in particular Schiffman encoding \cite{Schiffmann}, whose basic structure is a simple list of neurons with their connectivity information. This allows us to easily program our own rules for mutations to ensure that mutations do not result in illegal architectures (such as convolving to negative dimensions). Each vertex in the genome's graph encoding represents a layer in a CNN which stores hyper parameter information for the node it would construct, such as filter size, stride, padding, and the weight initialization method. 

To describe the adjusted NEAT algorithm, some basic mutations must be defined. \textit{Inject Node} injects a random node (convolution, pool, or ReLU) with pseudo-random hyperparameters into the genome's network between a preexisting connection. Before injecting the new node, we check to ensure that it would produce a valid network. If it does, the injection would occur, if it doesn't, we change the hyperparameters to values that would result in a valid network. This solves the problem of convolving the image to zero dimensions, as is guaranteed to occur as the number of injects increases. \textit{Inject Segment} injects a pair of a convolution layer with a ReLU activation with a pooling layer in a preexisting connection. \textit{Point Mutate} changes the essential hyperparameters of a node. These basic mutations allow for constructing most standard CNN architectures that one saw in literature before Inception-V1, such as AlexNet \cite{alexnet}. In addition to mutations, we incorporate speciation as outlined in \cite{NEAT} to allow for topological innovations to occur and prevent one topology framework from dominating the population.

\subsection{Dataset}
The dataset used for training and testing was compiled from the Prostate, Lung, Colorectal, and Ovarian Cancer Screening Trial (PLCO) dataset. This trial, organized by the National Cancer Institute, was a randomized, controlled trial to determine whether certain screening exams reduce the mortality of prostate, lung, colorectal, and ovarian cancer. Approximately 155,000 participants were enrolled in the screening portion of the trial from 1993 to 2006. If a participant developed cancer at any point during the screening phase, all CXRs preceding the diagnosis were considered to be cancerous and therefore were labelled positive, which resulted in about 4,000 positives in our dataset. We then selected the same amount of CXRs from patients who were never diagnosed with lung cancer, creating a balanced dataset. 

The dataset construction was designed to produce a model such that it could spot early stage lung cancer in CXR, as many of the positive labelled CXRs were taken years before the biopsy backed diagnosis occurred (which generally lead to a diagnosis of late stage lung cancer). Due to this construction, there are no annotations as to which areas of the CXR corresponded to the areas of the lungs that were biopsied, consequently preventing an accurate model from being used as a CADe tool due to lack of localization into the cancerous regions of the CXR. Additional work beyond an accurate classification model is required, which we explore in section 3.4

\begin{figure}[!b]
    \centering
    \includegraphics[]{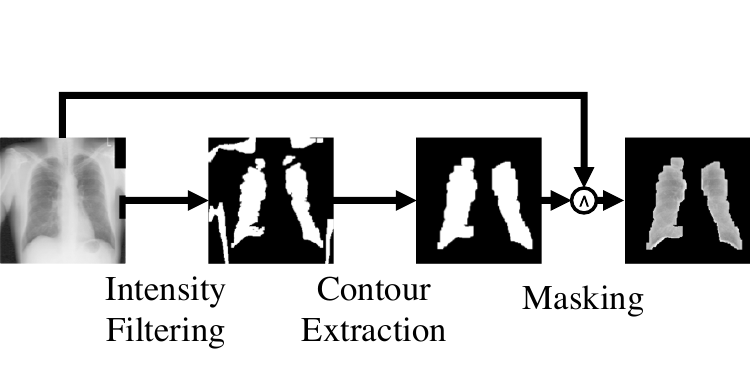}
    \caption{Stages of image preprocessing computer vision algorithm}
    \label{fig:cvflow}
\end{figure}

\textbf{Preprocessing}
The PLCO image dataset had roughly 200,000 CXRs with an approximate size of 2000x3000 pixels. In \cite{downscale}, down-scaling and cropping led to an improvement in their results, leading us to adopt similar procedures.

In the provided dataset, there were around 4,000 images that were linked with patients having biopsy-proven lung cancer. We used these images to create our cancer dataset, using an additional 4,000 other images as control/negative. 
Each of our datasets were then preprocessed to make the images uniform and to ensure the models were not influenced by other noise, such as notes or markings.

The first step was resampling and resizing the images to a constant 256x256 resolution, as this would retain enough detail for effective use and eliminate the inconsistencies seen in the input data.
The second step involved a specially-designed x-ray lung segmentation algorithm that was used to mask the input images to contain only the regions of interest. Initially, the images were shifted to have a constant mean, which was required due to the variation in images coming from different medical centers using different models of x-ray machines. Next, the images were filtered to retain only the middle intensities, which were found to correspond with the inside of the lungs. Contours were then extracted from the remaining part of the image which were then filled in to create the final mask for that image, which was then applied. The steps used in this procedure are as shown in figure \ref{fig:cvflow}.

Finally, the entire masked dataset as a whole was rescaled to a standard normal distribution which was then shuffled and randomly split into a training dataset containing 80\% of the final images, a validation dataset containing 10\%, and the last 10\% reserved for evaluating the final trained model.
\subsection{Training}
When training, we used a gene pool of 50 individuals, over the course of 10 generations. The genetic algorithm had mutation rate parameters that determined how often each mutation occurred. Each mutation received its own rate:
\begin{table}[H]
    \centering
    \begin{tabular}{ |p{3cm}||p{3cm}|}
        \hline
        Mutation & Rate \\
        \hline
        Inject Convolution & 50\% \\
        Inject Pooling & 50\% \\
        Add ReLU & 30\% \\
        Point Mutate & 45\% \\
        Inject Segment & 15\% \\
        \hline
    \end{tabular}
\end{table}
After the genetic algorithm produced a graph encoding of a network, we exported the graph into a Caffe model \cite{caffe}. Each model received identical training parameters, and they are as follows:
\begin{table}[H]
    \centering
    \begin{tabular}{ |p{3cm}||p{3cm}|}
        \hline
        Hyper Parameter & Value \\
        \hline
        Solver & SGD \\
        Epochs & 5 \\
        LR Policy & INV \\
        Base LR & 0.01 \\
        Power & .75 \\
        Momentum & .90 \\
        Weight Decay & 0.0005 \\
        \hline
    \end{tabular}
\end{table}
A chart detailing the population's best and average fitness over generations is shown in Figure 3. 

Because we used the error on the development set as a reward signal, it is important to confirm that the population was not over fitting on the validation set. We evaluated the best model produced by the genetic algorithm on the previously unseen test set and showed that the performance of the model was no worse than on the development set, confirming that the GA was not over fitting from the sparse, weak reward signal. The results of this evaluation is detailed in Table 1. 
\subsection{Results}
We ran several experiments with various population sizes and generations, and found that evolving 50 individuals for 10 generations was the most efficient in producing a fit population quickly as well as producing very fit top models. Unlike \cite{rl-nas1}, which trained 12,800 models to optimize the R-NN controller, we trained only 500 models.

\begin{figure}[!t]
    \centering
    \begin{tikzpicture}
        \selectcolormodel{gray}
        \begin{axis}[
            xlabel=Generation,
            ylabel=Accuracy (\%),
            xmin=0, xmax=10,
            ymin=0, ymax=100,
            clip=true,
            xticklabel style={
            /pgf/number format/fixed,
            /pgf/number format/precision=5
            },
            legend pos=south east,
            ytick={0,20,40,60,80,100},
            xtick={0,1,2,3,4,5,6,7,8,9,10},
            y filter/.code={\pgfmathparse{#1*100}\pgfmathresult}
            ]
        \addplot table [y=avg, x=gen]{fit.dat};
        \addlegendentry{Average}
        \addplot table [y=best, x=gen]{fit.dat};
        \addlegendentry{Best}
        \end{axis}
    \end{tikzpicture}
    \caption{Improvement of Accuracy over Generations}
    \label{fig:accuracy}
\end{figure}
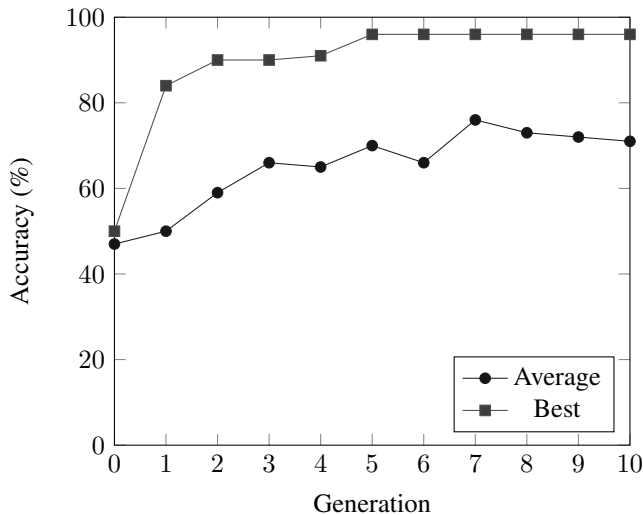

To test the effectiveness of DeepNEAT-Dx, we compared it to the best models from the past 7 years. As shown in the table above, DeepNEAT-Dx surpasses every model in accuracy by at least 5\%, and employed considerably fewer parameters. 

\begin{table}[!t]
    \centering
    \begin{tabular}{ |p{2.3cm}||p{1.5cm}|p{2.4cm}| }
        \hline
        Model & Accuracy & Parameter Count \\
        \hline
        AlexNet & 79.88\% & 62.3M\\
        Inception-V3 & 89.34\% & 23.8M\\
        ResNet-50 & 90.22\% & 25.6M\\
        ResNet-101 & 90.68\% & 44.5M\\
        VGG16 & 91.59\% & 138.3M \\
        ResNet-152 & 92.03\% & 60.2M\\
        \textbf{DeepNEAT-Dx} & \textbf{97.15\%} & \textbf{4.9M}\\
        \hline
    \end{tabular}
    \caption{Model comparison on CXR lung cancer dataset}
    \label{table:models}
\end{table}

\begin{table*}[!t]
    \centering
    \begin{tabular}{c|*{5}{>{\raggedright\arraybackslash}p{2.1cm}|}}
        \cline{3-4}
        \multicolumn{2}{r}{} & \multicolumn{2}{|c|}{ Predicted} & \multicolumn{2}{r}{} \\
        \cline{2-5}
         & Population: 1752 & Positive: 807 & Negative: 945 & Prevalence: 48.801\% & \multicolumn{1}{c}{} \\ \cline{1-6}
        \multicolumn{1}{|c|}{} & Positive: 855 & True Positive: 806 & False Negative: 49 & True Positive Rate: 94.27\% & False Negative Rate: 5.73\% \\  \cline{2-6}
        \multicolumn{1}{|c|}{\multirow{-2}{0.1cm}[5ex]{ \begin{sideways}Confirmed\end{sideways}}}
         & Negative: 897 & False Positive: 1 & True Negative: 896 & False Positive Rate: 0.11\% & True Negative Rate: 99.89\% \\ \hline
         & & Positive Predictive Value: 99.88\% & False Omission Rate: 5.19\% & Positive Likelihood Ratio: 845 & \\ \cline{3-5}
         & Accuracy: 97.15\% & False Discovery Rate: 0.12\% & Negative Predictive Value: 94.81\% & Negative Likelihood Ratio: 5.74\% & Diagnostic Odds Ratio: 14700 \\ \cline{2-6}
    \end{tabular}
    \vspace{0.2cm}
    \caption{Binary classifier evaluation contingency table and confusion matrix}
    \label{tab:matrix}
\end{table*}

\begin{figure*}[!t]
    \centering
\begin{tikzpicture}[thick,scale=0.55, every node/.style={scale=0.55}]
\tikzstyle{connection}=[ultra thick,every node/.style={sloped,allow upside down},draw=\edgecolor,opacity=0.7]
\tikzstyle{copyconnection}=[ultra thick,every node/.style={sloped,allow upside down},draw={rgb:blue,4;red,1;green,1;black,3},opacity=0.7]

\pic[shift={(0,0,0)}] at (0,0,0) 
    {Box={
        name=conv1,
        caption= ,
        xlabel={{32, }},
        zlabel=227,
        fill=\ConvColor,
        height=28,
        width=4,
        depth=28
        }
    };

\pic[shift={ (0,0,0) }] at (conv1-east) 
    {Box={
        name=pool1,
        caption= ,
        fill=\PoolColor,
        opacity=0.5,
        height=14,
        width=4,
        depth=14
        }
    };

\pic[shift={(1,0,0)}] at (pool1-east) 
    {Box={
        name=conv2,
        caption= ,
        xlabel={{64, }},
        zlabel=117,
        fill=\ConvColor,
        height=14,
        width=8,
        depth=14
        }
    };

\pic[shift={ (0,0,0) }] at (conv2-east) 
    {Box={
        name=pool2,
        caption= ,
        fill=\PoolColor,
        opacity=0.5,
        height=14,
        width=8,
        depth=14
        }
    };

\pic[shift={ (0,0,0) }] at (pool2-east) 
    {Box={
        name=pool3,
        caption= ,
        fill=\PoolColor,
        opacity=0.5,
        height=7,
        width=8,
        depth=7
        }
    };

\pic[shift={(1,0,0)}] at (pool3-east) 
    {Box={
        name=conv3,
        caption= ,
        xlabel={{128, }},
        zlabel=29,
        fill=\ConvColor,
        height=3,
        width=10,
        depth=3
        }
    };

\pic[shift={(1,0,0)}] at (conv3-east) 
    {Box={
        name=conv4,
        caption= ,
        xlabel={{128, }},
        zlabel=8,
        fill=\ConvColor,
        height=1,
        width=10,
        depth=1
        }
    };

\pic[shift={(1,0,0)}] at (conv4-east) 
    {Box={
        name=conv5,
        caption= ,
        xlabel={{512, }},
        zlabel=10,
        fill=\ConvColor,
        height=1,
        width=12,
        depth=1
        }
    };

\pic[shift={ (0,0,0) }] at (conv5-east) 
    {Box={
        name=pool4,
        caption= ,
        fill=\PoolColor,
        opacity=0.5,
        height=1,
        width=12,
        depth=1
        }
    };

\pic[shift={(1,0,0)}] at (pool4-east) 
    {Box={
        name=conv6,
        caption= ,
        xlabel={{64, }},
        zlabel=4,
        fill=\ConvColor,
        height=1,
        width=8,
        depth=1
        }
    };

\pic[shift={(3,0,0)}] at (conv6-east) 
    {Box={
        name=fc1,
        caption= ,
        xlabel={{" ","dummy"}},
        zlabel=1024,
        fill=\SoftmaxColor,
        opacity=0.8,
        height=3,
        width=1.5,
        depth=50
        }
    };

\pic[shift={(3,0,0)}] at (fc1-east) 
    {Box={
        name=fc2,
        caption= ,
        xlabel={{" ","dummy"}},
        zlabel=2,
        fill=\SoftmaxColor,
        opacity=0.8,
        height=3,
        width=1.5,
        depth=2
        }
    };

\draw [connection]  (pool1-east)    -- node {\midarrow} (conv2-west);

\draw [connection]  (pool3-east)    -- node {\midarrow} (conv3-west);

\draw [connection]  (conv3-east)    -- node {\midarrow} (conv4-west);

\draw [connection]  (conv4-east)    -- node {\midarrow} (conv5-west);

\draw [connection]  (pool4-east)    -- node {\midarrow} (conv6-west);

\draw [connection]  (conv6-east)    -- node {\midarrow} (fc1-west);

\draw [connection]  (fc1-east)    -- node {\midarrow} (fc2-west);

\end{tikzpicture}
\caption{Best model found by DeepNEAT algorithm after 10 generations}
    \label{fig:model}
\end{figure*}
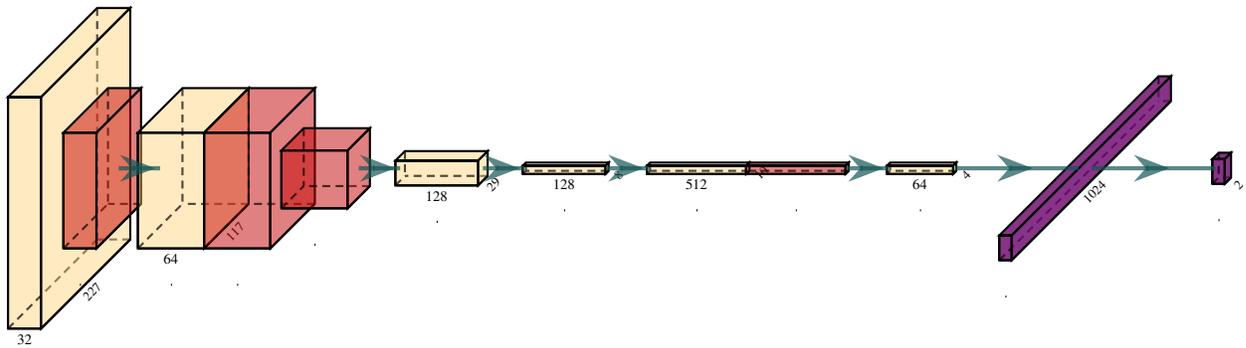

\begin{figure}[!t]
    \centering
    \begin{subfigure}[b]{0.1\textwidth}
        \centering
        \includegraphics[width=\textwidth]{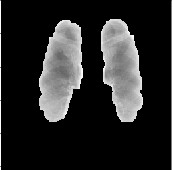}
        \caption{Input}
        \label{feat:input}
    \end{subfigure}
    \hfill
    \begin{subfigure}[b]{0.1\textwidth}
        \centering
        \includegraphics[width=\textwidth]{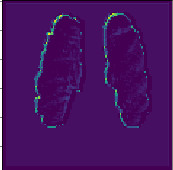}
        \caption{Feature 1}
        \label{feat:feat1}
    \end{subfigure}
    \hfill
    \begin{subfigure}[b]{0.1\textwidth}
        \centering
        \includegraphics[width=\textwidth]{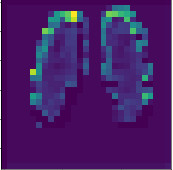}
        \caption{Feature 2}
        \label{feat:feat2}
    \end{subfigure}
    \hfill
    \begin{subfigure}[b]{0.1\textwidth}
        \centering
        \includegraphics[width=\textwidth]{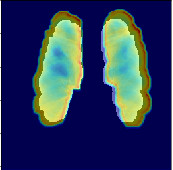}
        \caption{Heatmap}
        \label{feat:heat}
    \end{subfigure}
    \caption{Image with extracted features}
    \label{fig:feat}
\end{figure}

For binary classifiers, such as this model, a matrix can be used to present several statistics about the performance of the algorithm. The core of this matrix is the contingency table, which is used to present the frequency of the real condition variable and the predicted condition variable. From this cross-tabulation many ratios can be derived, most notably the rates of false positives and negatives among the population. To generate the contingency table in Table 1, the network was computed against all 1,884 validation samples and the ground truth label was recorded with the predicted output. The statistical ratios were then calculated from these inputs. The best model generated by our algorithm achieved an accuracy of 97.14\% with a PPV of 99.88\% and a NPV of 94.81\%.
\subsection{User Friendly Model Prediction Visualizations}

DeepNEAT-Dx produces a binary classifier, which is not sufficient enough to be allocated the label of being a medical CADe model. While an accurate binary classification of the presence of lung cancer within a particular CXR would help a radiologist in deciding what further testing to do for the patient or how to treat the patient, its lack of a useful localization of the classification dampens the impact of the model. Because of the general lack of annotations in the dataset, it is impossible to deploy a traditional object detection model, forcing the use of advanced visualization techniques.

Recent attempts to demystify the neural network black box have included the Grad-CAM \cite{gradcam} project, which introduced a technique to visualize the weights of a layer that most strongly correlate with the final decision. It attempts to maximize the outputs of that particular layer with respect to a desired filter (class). This technique was used to generate the heatmaps representing sources of influence to the prediction shown in figure \ref{fig:sample} and figure \ref{feat:heat}, where the selected layer was the final fully connected layer in the graph from the most recent convolution layer and the filter was for images which were predicted to be in the cancerous class. It is important to note that it is impossible to verify whether the heatmap activations correspond to the actual location of the malignant nodules or masses given the binary dataset. Coincidentally, the PLCO dataset provided radiologist annotations detailing a variety of lung abnormalities and their locations. One such abnormality was the presence of nodule masses. The heatmap markings seen in Figure \ref{fig:sample} correspond directly to the nodule masses noted location by a radiologist. However, it is important to clarify that a nodule mass does not always lead to lung cancer, but it is an interesting correlation. 
\section{Discussion}

The final accuracy of the network over the test set of 97.15\% shows that the model has succeeded at learning features related to the presence of various types of confirmed lung cancer in these images. Given that a human radiologist must spend a considerable amount of time on each image to make a correct prediction, screenings for many types of cancer do not often occur early, causing diagnoses to often arrive during the late stages of these diseases. The potential to use such a model as a preliminary screening step may save many lives with early detection and from misdiagnoses. All cases of concern will necessarily have to be verified by an experienced radiologist, but the automation provided by this tool will decrease costs, increase the accessibility of screenings, and increase the speed and accuracy of diagnoses.


{\small
\bibliographystyle{ieee}
\bibliography{research}

\begin{thebibliography}{10}\itemsep=-1pt

\bibitem{lungcancer}
V.~Ambrosini, S.~Nicolini, P.~Caroli, C.~Nanni, A.~Massaro, M.~C. Marzola,
  D.~Rubello, and S.~Fanti.
\newblock Pet/ct imaging in different types of lung cancer: An overview.
\newblock {\em European Journal of Radiology}, pages 998--1001, 2012.

\bibitem{learning2learn}
M.~Andrychowicz, M.~Denil, S.~G\'{o}mez, M.~W. Hoffman, D.~Pfau, T.~Schaul,
  B.~Shillingford, and N.~de~Freitas.
\newblock Learning to learn by gradient descent by gradient descent.
\newblock In D.~D. Lee, M.~Sugiyama, U.~V. Luxburg, I.~Guyon, and R.~Garnett,
  editors, {\em Advances in Neural Information Processing Systems 29}, pages
  3981--3989. Curran Associates, Inc., 2016.

\bibitem{rnnevolve}
P.~J. Angeline, G.~M. Saunders, and J.~B. Pollack.
\newblock An evolutionary algorithm that constructs recurrent neural networks.
\newblock {\em IEEE Transactions on Neural Networks}, 5:54--65, 1994.

\bibitem{cnnlungcancer}
M.~Anthimopoulos, S.~Christodoulidis, L.~Ebner, A.~Christe, and S.~Mougiakakou.
\newblock Lung pattern classification for interstitial lung diseases using a
  deep convolutional neural network.
\newblock {\em IEEE Transactions on Medical Imaging}, 35(5):1207--1216, 5 2016.

\bibitem{encoding2}
E.~G. {Carrano}, C.~M. {Fonseca}, R.~H.~C. {Takahashi}, L.~C.~A. {Pimenta}, and
  O.~M. {Neto}.
\newblock A preliminary comparison of tree encoding schemes for evolutionary
  algorithms.
\newblock In {\em 2007 IEEE International Conference on Systems, Man and
  Cybernetics}, pages 1969--1974, Oct 2007.

\bibitem{nngenalg}
D.~Dasgupta and D.~McGregor.
\newblock Designing application-specific neural networks using the structured
  genetic algorithm.
\newblock IEEE, 1992.

\bibitem{skincancer}
A.~Esteva, B.~Kuprel, R.~A. Novoa, J.~Ko, S.~M. Swetter, H.~M. Blau, and
  S.~Thrun.
\newblock Dermatologist-level classification of skin cancer with deep neural
  networks.
\newblock {\em Nature}, pages 115--118, 2017.

\bibitem{cadbreastcancer}
T.~Freer and M.~J. Ulissey.
\newblock Screening mammography with computer-aided detection: prospective
  study of 12,860 patients in a community breast center.
\newblock {\em Radiology}, 220 3:781--6, 2001.

\bibitem{downscale}
L.~G. Hafemann, L.~S. Oliveira, and P.~Cavalin.
\newblock Forest species recognition using deep convolutional neural networks.
\newblock In {\em 2014 22nd International Conference on Pattern Recognition},
  pages 1103--1107, 9 2014.

\bibitem{microsoft2016}
K.~He, X.~Zhang, S.~Ren, and J.~Sun.
\newblock Deep residual learning for image recognition.
\newblock Computer Vision and Pattern Recognition, 2016.

\bibitem{caffe}
Y.~Jia, E.~Shelhamer, J.~Donahue, S.~Karayev, J.~Long, R.~Girshick,
  S.~Guadarrama, and T.~Darrell.
\newblock Caffe: Convolutional architecture for fast feature embedding.
\newblock In {\em Proceedings of the 22Nd ACM International Conference on
  Multimedia}, MM '14, pages 675--678, New York, NY, USA, 2014. ACM.

\bibitem{eff-nas}
H.~Jin, Q.~Song, and X.~Hu.
\newblock Efficient neural architecture search with network morphism.
\newblock {\em CoRR}, abs/1806.10282, 2018.

\bibitem{adam}
D.~Kingma and J.~Ba.
\newblock Adam: A method for stochastic optimization.
\newblock {\em International Conference on Learning Representations}, 12 2014.

\bibitem{alexnet}
A.~Krizhevsky, I.~Sutskever, and G.~E. Hinton.
\newblock Imagenet classification with deep convolutional neural networks.
\newblock {\em Neural Information Processing Systems}, 2012.

\bibitem{yann1998}
Y.~LeCun, L.~Bottou, Y.~Bengio, and P.~Haffner.
\newblock Gradient-based learning applied to document recognition.
\newblock {\em Proceedings of the IEEE}, Vol. 86, Issue 11:2278--2324, 1998.

\bibitem{prog-nas}
C.~Liu, B.~Zoph, J.~Shlens, W.~Hua, L.~Li, L.~Fei{-}Fei, A.~L. Yuille,
  J.~Huang, and K.~Murphy.
\newblock Progressive neural architecture search.
\newblock {\em CoRR}, abs/1712.00559, 2017.

\bibitem{diff-nas}
H.~Liu, K.~Simonyan, and Y.~Yang.
\newblock {DARTS:} differentiable architecture search.
\newblock {\em CoRR}, abs/1806.09055, 2018.

\bibitem{deepcovnets}
S.~{Liu} and W.~{Deng}.
\newblock Very deep convolutional neural network based image classification
  using small training sample size.
\newblock In {\em 2015 3rd IAPR Asian Conference on Pattern Recognition
  (ACPR)}, pages 730--734, Nov 2015.

\bibitem{NAS}
Z.~Lu, I.~Whalen, V.~Boddeti, Y.~Dhebar, K.~Deb, E.~Goodman, and W.~Banzhaf.
\newblock Nsga-net: neural architecture search using multi-objective genetic
  algorithm.
\newblock pages 419--427, 07 2019.

\bibitem{deepneat}
R.~Miikkulainen, J.~Z. Liang, E.~Meyerson, A.~Rawal, D.~Fink, O.~Francon,
  B.~Raju, H.~Shahrzad, A.~Navruzyan, N.~Duffy, and B.~Hodjat.
\newblock Evolving deep neural networks.
\newblock {\em CoRR}, abs/1703.00548, 2017.

\bibitem{dualrep}
J.~C.~F. Pujol and R.~Poli.
\newblock Evolving the topology and the weights of neural networks using a dual
  representation, 1998.

\bibitem{GA-RL}
T.~Salimans, J.~Ho, X.~Chen, and I.~Sutskever.
\newblock Evolution strategies as a scalable alternative to reinforcement
  learning.
\newblock 03 2017.

\bibitem{Schiffmann}
W.~Schiffmann, M.~Joost, and R.~Werner.
\newblock Performance evaluation of evolutionary created neural network
  topologies.
\newblock {\em Springer-Verlag London}, Vol. 2:274--283, 1990.

\bibitem{gradcam}
R.~R. {Selvaraju}, M.~{Cogswell}, A.~{Das}, R.~{Vedantam}, D.~{Parikh}, and
  D.~{Batra}.
\newblock Grad-cam: Visual explanations from deep networks via gradient-based
  localization.
\newblock In {\em 2017 IEEE International Conference on Computer Vision
  (ICCV)}, pages 618--626, Oct 2017.

\bibitem{parallelGA}
R.~Shonkwiler.
\newblock Parallel genetic algorithms.
\newblock In {\em Proceedings of the 5th International Conference on Genetic
  Algorithms}, pages 199--205, San Francisco, CA, USA, 1993. Morgan Kaufmann
  Publishers Inc.

\bibitem{lungcancersurvey}
R.~L. Siegel, K.~D. Miller, and A.~Jemal.
\newblock Cancer statistics, 2019.
\newblock {\em CA: A Cancer Journal for Clinicians}, 69(1):7--34, 2019.

\bibitem{breastcancercnns}
F.~A. Spanhol, L.~S. Oliveira, C.~Petitjean, and L.~Heutte.
\newblock Breast cancer histopathological image classification using
  convolutional neural networks.
\newblock In {\em 2016 International Joint Conference on Neural Networks
  (IJCNN)}, pages 2560--2567, 8 2016.

\bibitem{breakhis}
F.~A. Spanhol, L.~S. Oliveira, C.~Petitjean, and L.~Heutte.
\newblock A dataset for breast cancer histopathological image classification.
\newblock {\em IEEE Transactions on Biomedical Engineering}, 63(7):1455--1462,
  8 2016.

\bibitem{dropout2012}
N.~Srivastava, G.~Hinton, A.~Krizhevsky, I.~Sutskever, and R.~Salakhutdinov.
\newblock Dropout: A simple way to prevent neural networks from overfitting.
\newblock {\em Journal of Machine Learning Research}, 15:1929--1958, 2014.

\bibitem{NEAT}
K.~O. Stanley and R.~Miikkulainen.
\newblock Evolving neural networks through augmenting topologies.
\newblock {\em Journal of Evolutionary Computation}, Vol. 10 Issue 2:99--127,
  2002.

\bibitem{googlenet2015}
C.~Szegedy, W.~Liu, Y.~Jia, S.~Reed, D.~Anguelov, D.~Erhan, V.~Vanhouck, and
  A.~Rabinovich.
\newblock Going deeper with convolutions.
\newblock {\em Computer Vision and Pattern Recognition}, 2015.

\bibitem{cad-survey}
R.~Takahashi and Y.~Kajikawa.
\newblock Computer-aided diagnosis: A survey with bibliometric analysis.
\newblock {\em International Journal of Medical Informatics}, 101:58 -- 67,
  2017.

\bibitem{efficientnet}
M.~Tan and Q.~Le.
\newblock {E}fficient{N}et: Rethinking model scaling for convolutional neural
  networks.
\newblock In K.~Chaudhuri and R.~Salakhutdinov, editors, {\em Proceedings of
  the 36th International Conference on Machine Learning}, volume~97 of {\em
  Proceedings of Machine Learning Research}, pages 6105--6114, Long Beach,
  California, USA, 09--15 Jun 2019. PMLR.

\bibitem{overlooked}
N.~Wu, G.~Gamsu, J.~Czum, B.~Held, R.~Thakur, and G.~Nicola.
\newblock Detection of small pulmonary nodules using direct digital radiography
  and picture archiving and communication systems.
\newblock {\em Journal of thoracic imaging}, 21:27--31, 04 2006.

\bibitem{rl-nas2}
Z.~{Zhong}, J.~{Yan}, W.~{Wu}, J.~{Shao}, and C.~{Liu}.
\newblock Practical block-wise neural network architecture generation.
\newblock In {\em 2018 IEEE/CVF Conference on Computer Vision and Pattern
  Recognition}, pages 2423--2432, June 2018.

\bibitem{rl-nas1}
B.~Zoph and Q.~V. Le.
\newblock Neural architecture search with reinforcement learning.
\newblock {\em CoRR}, abs/1611.01578, 2016.

\end{thebibliography}
}

\end{document}